\documentclass[conference]{IEEEtran}
\IEEEoverridecommandlockouts
\usepackage{graphicx}
\usepackage{color}
\usepackage{chngpage}
\usepackage{multicol}
\usepackage{array,multirow}
\usepackage{amssymb}
\usepackage{amsmath}
\usepackage{bm}
\usepackage{subcaption}
\usepackage{tabularx,booktabs}
\usepackage[bookmarks=false]{hyperref}
\newcolumntype{Y}{>{\centering\arraybackslash}X}

\usepackage[utf8]{inputenc}

\usepackage{xcolor}
\definecolor{RoyalBlue}{RGB}{65, 105, 225}

\newcolumntype{?}{!{\vrule width 1pt}}

\def\BibTeX{{\rm B\kern-.05em{\sc i\kern-.025em b}\kern-.08em
    T\kern-.1667em\lower.7ex\hbox{E}\kern-.125emX}}

\begin{document}

\title{
2D Convolutional Neural Networks for 3D Digital Breast Tomosynthesis Classification
\thanks{This study is sponsored by Grant No. IRG 16-182-28 from the American Cancer Society and Grant No. IIS-1553116 from the National Science Foundation.}
}
\author{\textbf{Yu~Zhang\textsuperscript{1}*,
        Xiaoqin~Wang\textsuperscript{2}
		Hunter~Blanton\textsuperscript{1},
		Gongbo~Liang\textsuperscript{1},
		Xin~Xing\textsuperscript{1},
		Nathan~Jacobs\textsuperscript{1}
		 } \\ [1ex]
		$1$ Department of Computer Science, University of Kentucky, Lexington, KY, USA\\
		$2$ Department of Radiology, University of Kentucky, Lexington, KY, USA \\
		Email: y.zhang@uky.edu*}

\maketitle

\begin{abstract}
Automated methods for breast cancer detection have focused on 2D mammography and have largely ignored 3D digital breast tomosynthesis (DBT), which is frequently used in clinical practice. The two key challenges in developing automated methods for DBT classification are handling the variable number of slices and retaining slice-to-slice changes. We propose a novel deep 2D convolutional neural network (CNN) architecture for DBT classification that simultaneously overcomes both challenges. Our approach operates on the full volume, regardless of the number of slices, and allows the use of pre-trained 2D CNNs for feature extraction, which is important given the limited amount of annotated training data.  In an extensive evaluation on a real-world clinical dataset, our approach achieves $0.854$ auROC, which is $28.80\%$ higher than approaches based on 3D CNNs. We also find that these improvements are stable across a range of model configurations.
\end{abstract}


\section{Introduction}

Recent global burden of disease study reported that breast cancer is the most common cancer type and is a leading cause of cancer-induced mortality among women with 2.4 million new cases and 523,000 deaths per year around the world. Detection of breast cancer in early stages reduces the death rate and 2D X-ray mammography is commonly used in clinic for early detection. However, 2D mammogram suffers from a significant high false-positive recalls and the sensitivity of cancer detection is also limited by superimposition of the breast tissue. 

A newly emerging technology: 3D full-volume digital breast tomosynthesis (DBT), has been proven to improve cancer detection as it can reduce the masking effect of the breast tissue on 2D mammogram by creating multiple 2D imaging slices of the breast. There is increasing utilization of this new imaging modality in practice. However, the interpretation of DBT is time consuming and requires additional training. This demand can limit the clinical utilization of this promising technology. Therefore, it is important to develop automated DBT interpretation tools which can aid radiologists to detect breast cancer more accurately and efficiently so that this potentially life-saving technology can be applied more broadly and benefit more patients. 

\begin{figure}[!tb]
\centering
\includegraphics[width=0.49\textwidth]{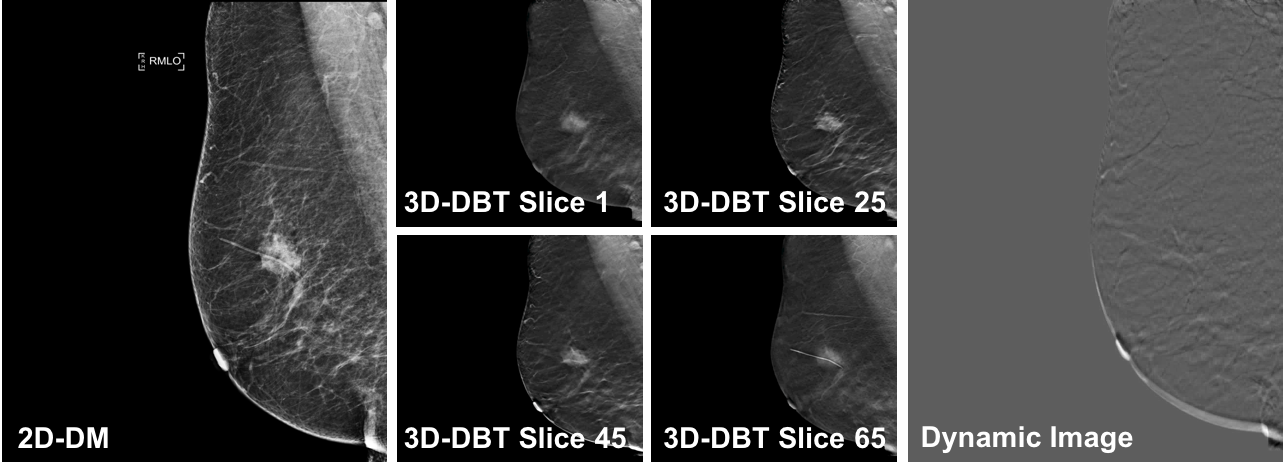}
\caption{Example of a negative 2D digital mammogram in the mediolateral oblique (MLO) view (left), the four slices from the associated 3D digital breast tomosynthesis (middle), and generated dynamic image (right).}
\label{fig:mammo}
\end{figure}

With the recent rapid development of deep learning, new approaches with deep convolutional neural networks (CNNs) have shown great promise in increasing the classification accuracy in mammogram. However, most published work on automated breast cancer detection have focused on 2D mammogram~\cite{zhang2017whole}~\cite{arevalo2016representation}, but very few focus on exploring deep learning models on DBT because of three main challenges: high computational requirements that come with large sizes of DBT, lack of available pre-trained 3D CNN models, and lack of public DBT dataset.

In this paper, we propose a novel approach for full-volume DBT classification that enables us to use pre-trained 2D CNNs and significantly improves the classification performance. We compare the effect of different pooling methods, feature extractors, and data fusion strategies on model performance. By taking advantage of 2D CNNs, we reduce the computational requirements of exploring DBT in the deep learning manner and avoid the need of 3D CNN pre-trained models. We evaluate our models on a private clinical dataset. Every patient case in the dataset contains both full-field 2D mammogram and full-volume DBT, allowing for comparison between methods using either modality. Figure~\ref{fig:mammo} shows a negative sample including a 2D mammogram, four DBT slices, and a dynamic image generated from DBT. Instead of  focusing on the classification of small tumor patches in most of the previous work~\cite{hua2015computer}~\cite{dhungel2017fully}, our model focuses on full-volume classification.

\begin{figure*}
  \centering
  \includegraphics[width=1.0\textwidth]{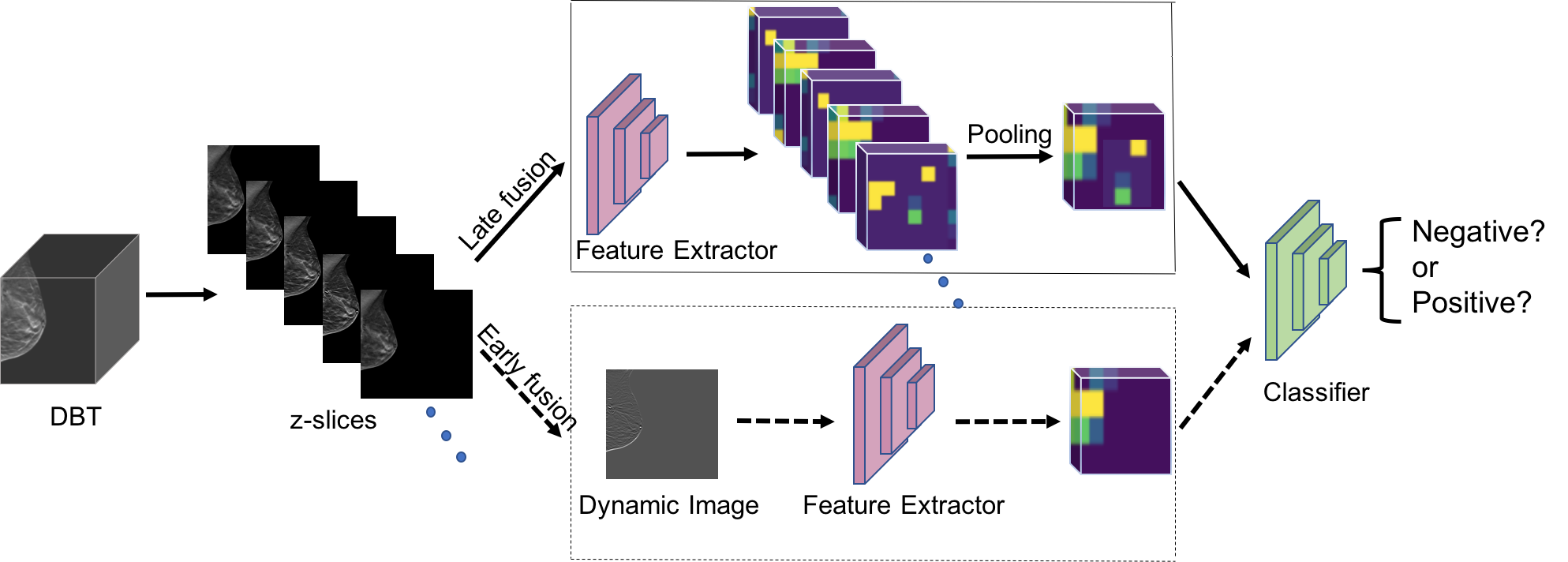}
  \caption{The Architecture of our pipeline. Each DBT video is represented by a sequence of slices. The late fusion branch (upper) and the early fusion branch (lower) illustrate different data fusion strategies used in our work.}
  \label{figs:flow}
\end{figure*}

In summary, the proposed approach has the following contributions:
\begin{itemize}
\item We propose a novel approach for DBT classification that enables us to use pre-trained 2D CNNs and significantly improves the performance (from 0.663 auROC to 0.854 auROC).
\item We compare the effects of different pooling methods, feature extractors, and data fusion strategies.
\item We focus on full-volume classification instead of only using small tumor patches and discuss the performance of space-to-channel encoding.

\end{itemize}

\section{Approach}

In this section, we introduce the architecture for automated DBT classification, the loss function used, and the implementation details.

\subsection{Architecture Overview}
Our base convolutional neural network model (Figure~\ref{figs:flow}) takes as input a three-dimentional digital breast tomosynthesis (a sequence of slices, which we named z-slices in the following), and outputs a binary distribution over a discrete label space (negative or positive). Expanding on previous work~\cite{whole}, we leverage AlexNet~\cite{krizhevsky2012imagenet} pre-trained on ImageNet~\cite{deng2009imagenet} for feature extraction to train a small classification network. However, unlike 2D mammogram, DBT is variable in size. To overcome this, features are extracted for each set of z-slices and consolidated into a fixed size feature map by pooling across the channel dimension. Final classification is performed on the output feature map using a separate network trained from scratch for our task. The above approach is named \textit{late fusion}, illustrated in Figure~\ref{figs:flow} (upper branch).

Besides this late fusion approach, we also explore some other strategies to fuse the full sequence of slices at the early point in the pipeline, named \textit{early fusion} in the lower branch in Figure~\ref{figs:flow}. We explore two different methods for this early fusion strategy.

The first method we try is averaging all z-slices and generate a single pseudo 2D mammogram for the feature extractor, then the classifier takes the feature map as the only input and outputs the predicted label.

The second method is using the Dynamic Image Networks, which input the z-slices and output a dynamic image, to fuse the sequence of images. Following the previous work~\cite{bilen2017action}, our dynamic image is obtained by a ranking function $f$ with parameters $\psi$ for a sequence of slices $I_1$,...,$I_T$.
\begin{equation}
  \begin{array}{l}
      I_{dynamic} = f(I_1,...,I_T;\psi)
  \end{array}
\end{equation}
The slices are directly mapped to a function, and we directly compute the dynamic image $I_{dynamic}$. Then the feature extractor extracts a feature map from the computed dynamic image, and use this feature map directly for classification. Early fusion branch in Figure~\ref{figs:flow} illustrates this process.

\subsection{Loss Function}
The classifier is trained using binary cross-entropy loss. For every feature map $m$, the classifier predicts the probability $C(z) = P(l=0|m)$ that the $m$ is labeled as negative ($l=0$). The loss function is showed in \eqref{loss}.
\begin{equation}
\begin{split}
L_{classifier}(C(z), t) = & -\dfrac{1}{n}\sum_{i=1}^{n}[t^{i}\log({C(z)^{i}})+ \\
&(1-t^{i})\log(1-{C(z)^{i}})]. \label{loss}
\end{split}
\end{equation}
For negative samples, we use $l=0$, and $l=1$ for positive samples.

\begin{table*}[ht]
\caption{Evaluation results of our approaches trained on different settings vs. previous 3D CNN based work~\cite{whole}.
}
\centering
\begin{tabular}{ccccccccc}
\hline
Approach & Architecture & Fusion Method & Pooling Method & Batch Size & Learning Rate & Dropout & auROC\\
\hline
3D-A1~\cite{whole} & 3 Conv. layers & - & max pooling & 128 & 0.01 & 0.5  & 0.631\\
3D-T1-Alex~\cite{whole} & 3D CNN+AlexNet &- & max pooling & 16 & 0.01 & 0.5  & 0.612\\
3D-T2-Alex~\cite{whole} & 3D CNN+AlexNet &- & max pooling & 16 & 0.0001 & 0.5  & 0.663\\ \hline

Ours (min pooling) & AlexNet & late & min pooling & 256 & 0.0001 & 0.5  & 0.789\\
Ours (avg. pooling) & AlexNet & late & avg. pooling & 256 & 0.0001 & 0.5  & 0.823\\
Ours (max pooling) & AlexNet & late & max pooling & 256 & 0.0001 & 0.5  &\textbf{0.854}\\

Ours (ResNet) & ResNet & late & max pooling & 256 & 0.0001 & 0.5  & 0.841\\
Ours (Xception) & Xception & late & max pooling & 256 & 0.0001 & 0.5  & 0.830\\

Ours (dynamic image) & AlexNet & early (dynamic) & max pooling & 256 & 0.0001 & 0.5  & 0.761\\
Ours (average image) & AlexNet & early (average) & max pooling & 256 & 0.0001 & 0.5  & 0.792\\

\hline
\end{tabular}
\label{tab:pooling}
\end{table*}

\begin{figure}
  \includegraphics[width=\linewidth]{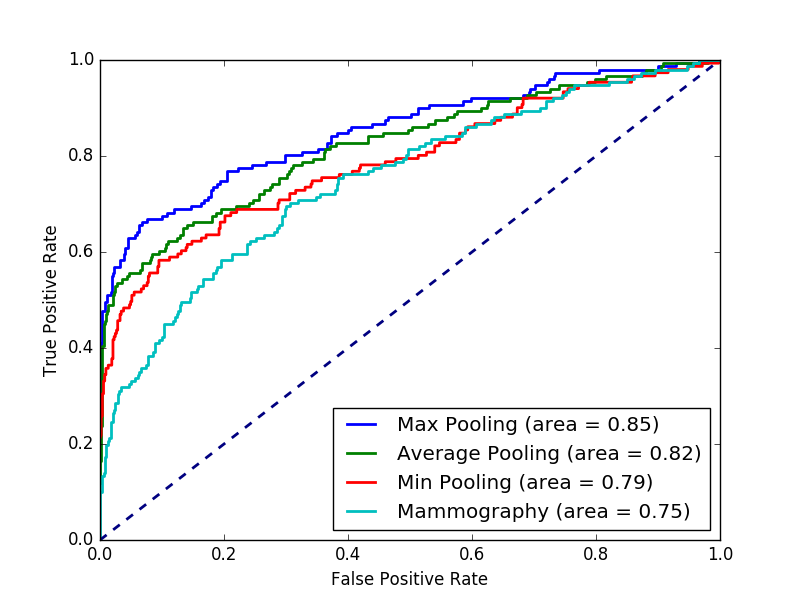}
  \caption{Comparison among different pooling methods and max pooling on 2D mammography. The curves show that all pooling methods on DBT outperform using 2D mammography.}
  \label{fig:pooling}
\end{figure}

\subsection{Implementation Details}
Our approach was implemented in Pytorch and trained and tested on a Ubuntu server with two Nvidia GTX $1080$ GPUs. We test three different networks as the feature extractor, AlexNet~\cite{krizhevsky2012imagenet}, ResNet50~\cite{b3}, and Xception~\cite{chollet2017xception}. For the classifier, we use a combination of two linear layers attached to a convolutional layer. We use Adam as our optimizer. L2 regularization and dropout are also used to prevent the networks from overfitting.

\section{Evaluation}
In this section, we evaluate multiple architectures with different training settings. Our results (in Table~\ref{tab:pooling}) show the combination of \textit{AlexNet}, \textit{max pooling}, and \textit{late fusion strategy} achieves the best performance $0.854$ auROC, which is $28.80\%$ higher than the best performance of the baseline method, which is only $0.663$ auROC.

\begin{figure}
  \includegraphics[width=\linewidth]{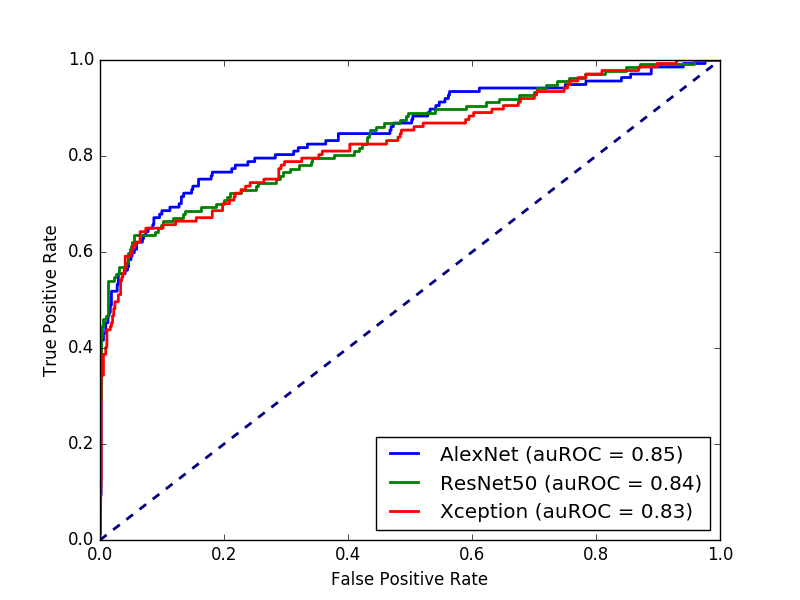}
  \caption{Comparison among AlexNet, ResNet50, and Xception. The curves show that AlexNet achieves the highest auROC.}
  \label{fig:nets}
\end{figure}
\subsection{Dataset}
A private clinical dataset is used in this study. All the 2D mammogram and DBT data was retrospectively collected from patients seen at a comprehensive breast cancer center between 2014 and 2017. 
This study was reviewed by the Institutional Review Board and was compliant with the Health Insurance Portability and Accountability Act. 

The dataset we use contains $3018$ negative and $272$ malignant samples. This is consistent with the previous work~\cite{whole}. We also use another $415$ benign samples for a ternary classification. Both the benign and malignant cases were proved with biopsy. All cases had both 2D mammogram and DBT in either craniocaudal (CC) or mediolateral oblique (MLO) view or both views. Approximately $1400$ paired DM/DBT data were included. To the best of our knowledge, this is currently the largest paired DM/DBT breast cancer dataset. We split the data in $80\%$ for training and $20\%$ for testing. For binary negative/positive classification task in this work, only malignant cases are labeled as positive, and this is done to be consistent with previous work~\cite{whole}. The original DBT z-slice resolution of $1024 \times 1024$ is preserved in all experiments.

\subsection{Balanced Sampling}

Due to the very unbalanced labels in the dataset, as the negative class is over 7 times as large as the benign class and over 4 times as large as the malignant class, naive training may lead the network to poor fitting to negative predictions. To alleviate this issue, each mini-batch is created in using balanced sampling which is the same as the previous work~\cite{whole}. To aid in generality, data augmentation was performed with random horizontal flips and rotations by 0, 90, 180, or 270 degrees, effectively increased our training set size by a factor of 8. 

\begin{figure*}
  \centering
  \includegraphics[width=1.0\textwidth]{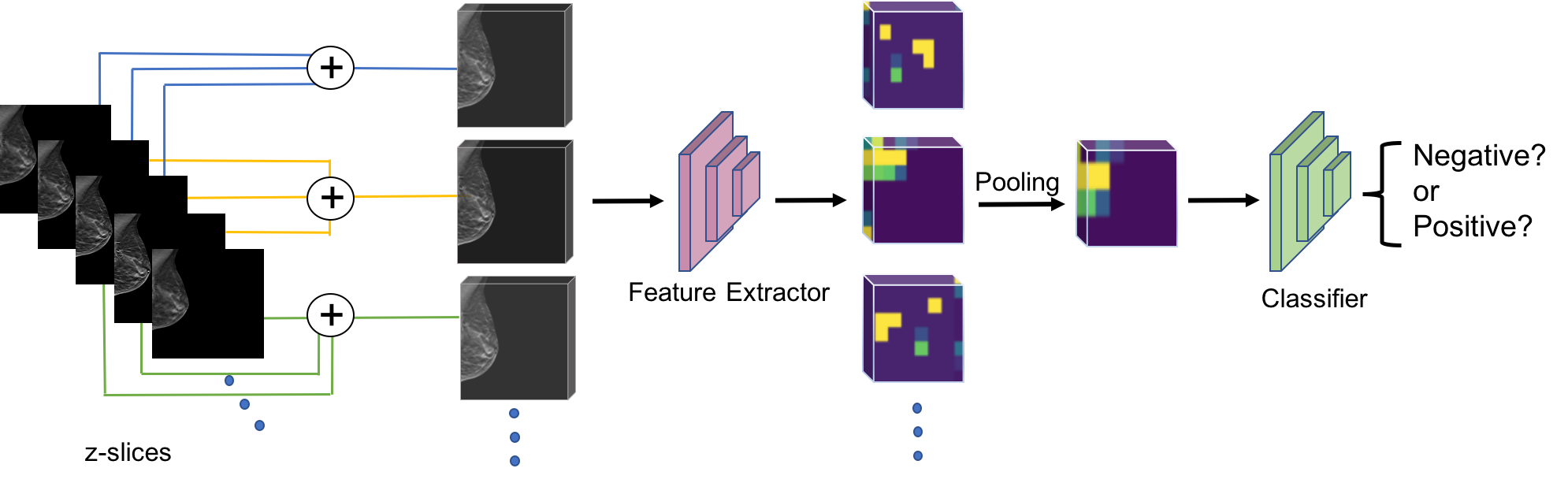}
  \caption{Space-to-channel encoding ($j=1$). Every 3 consecutive z-slices are stacked as a 3 channel image input to the feature extractor. Then generate a feature map by pooling the features extracted for the binary classification.}
  \label{figs:consistency}
\end{figure*}

\subsection{Feature-wise Pooling}
The fundamental idea behind our method is the pooling of features along a variable number of depth-wise computed feature maps. We test three types of pooling, minimum pooling, average pooling, and maximum pooling on DBT. Cross validation is used and the average testing results as well as training parameters are shown in Table~\ref{tab:pooling}. The results show that our approach w/ max pooling achieves the best performance, which is $0.854$ auROC (Area Under the Receiver Operating Characteristics). This is similar to many other 3D tasks where gobal max pooling was found to outperform average and minimum pooling. The best performance achieved by the previous 3D CNN approaches~\cite{whole} is $0.663$ auROC, which is lower than all three pooling methods of ours.

It is important to note that our approach outperforms using 2D mammography, which suggests that 2D CNNs can classify the breast tumor on DBT better than on 2D mammogram. Figure~\ref{fig:pooling} shows the comparison of ROC curves.

\subsection{Choice of Feature Extractor}
The feature extractor is a vital component of our system and its usefulness for this task needs to be determined. We test three common classification networks, AlexNet, ResNet50, and Xception. Each pretrained on ImageNet. Features are extracted just before the fully connected networks. Given the input size of $1024 \times 1024$, these networks given varying output feature map shapes. Results are shown for binary classification between negative (no tumor) and positive (malignant tumor).

\begin{itemize}
    \item ResNet50 : $4 \times 4 \times 2048$ (32768 elements) 
    \item AlexNet : $31 \times 31 \times 256$ (246016 elements)
    \item Xception : $32 \times 32 \times 2048$ (2097152 elements)
\end{itemize}

The varying feature map sizes, both spatially and in total number of elements, provide possible insight into the importance of spatial resolution and number of channels of the feature maps. The results of using each of these networks for feature extraction are shown in Table~\ref{tab:pooling} and the ROC curves are shown in Figure~\ref{fig:nets}. We notice that AlexNet achieves outperforms other architectures, though the difference is not significant and can be attributed to randomness in the training process. For this reason, we choose AlexNet for continued experiments for its simplicity and to maintain consistency with previous work~\cite{whole} for fair comparison.

\subsection{Where to Fuse Z-Slices}
\begin{figure}
  \includegraphics[width=\linewidth]{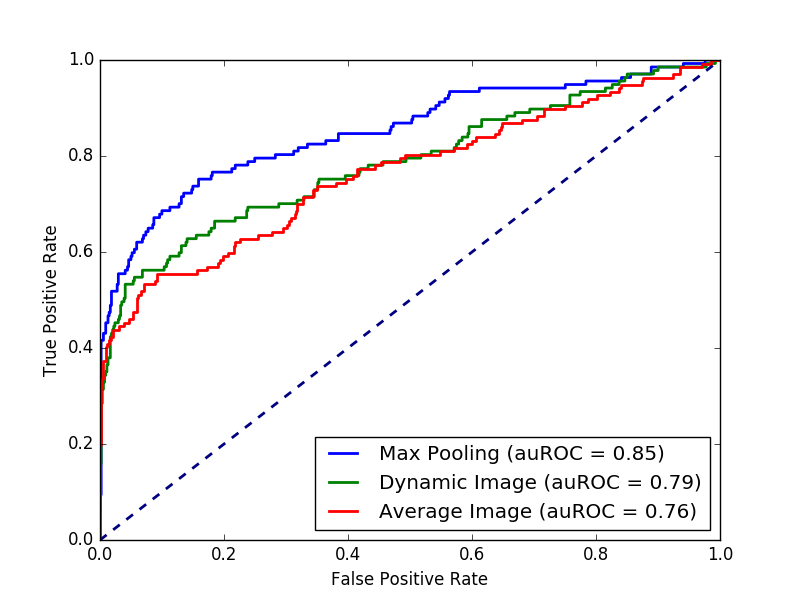}
  \caption{Comparison among different fusion strategies: max pooling for late fusion, dynamic image and average image for early fusion. Late fusion (w/ max pooling) achieves the highest auROC $0.85$.}
  \label{fig:fusion}
\end{figure}

Beyond choice of the pooling method and the feature extractor, it is importand to explore where in the pipeline to fuse the sequence of images is the best. We evaluate late fusion and early fusion. For late fusion, we extract features first, and then generate a single feature map by pooling. For early fusion, we explore two different strategies. One is averaging all slices to get a psudo 2D mammogram, the other is calculating a dynamic image based on the sequence of z-slices. We only show max pooling for late fusion since it has been proven to be the best pooling method. We evaluate all fusion strategies and show the results in Table~\ref{tab:pooling} and ROC curves in Figure~\ref{fig:fusion}. In general, our results show that late fusion strategies outperform early fusion strategies. However, for the early fusion itself, using dynamic image achieves $0.792$ auROC, which is better than using average image ($0.761$ auROC). That supports the idea that slice-to-slice information acquired by using dynamic image is useful for DBT classification tasks.

\subsection{Space-to-Channel Encoding}

The performance of using dynamic image inspires us to explore the slice-to-slice correspondence of DBT. The value of DBT is the ability to visualize anomalies as they manifest along consecutive z-slices of tissue. However, this benefit is ignored in the pooling process described above. By training a classifier on the pooling features among all z-slices, we are treating the 3D volume as a random set of z-slices instead of a physically consistent sequence of tissue slices. As it is the slice-to-slice correspondence that is typically useful during clinical examination, it is possible that this correspondence information would be useful for a classifier.

We encode the DBT slices from space to channel to take advantage this slice-to-slice correspondence. In more details, we use 3 unique z-slices of the volume as 3 channel input to the feature extractor. In general, the equation for slice selection becomes:
\begin{equation}
I = \{V_{i-j}, V_{i}, V_{i+j}\} \quad \forall \ i\in\{1,N\}. \label{selection}
\end{equation}
This results in a 3 channel image centered around slice $i$, using surrounding z-slices to embed information from the 3rd dimension. For example, $j=0$ results in the approach above, $j=1$ would use 3 consecutive z-slices, $j=2$ would skip a single z-slice on each side. As j increases, the similarity between z-slices decreases, but the effective spatial extend increases. Figure~\ref{figs:consistency} illustrates the process when $j=1$. 

\begin{figure}
  \includegraphics[width=\linewidth]{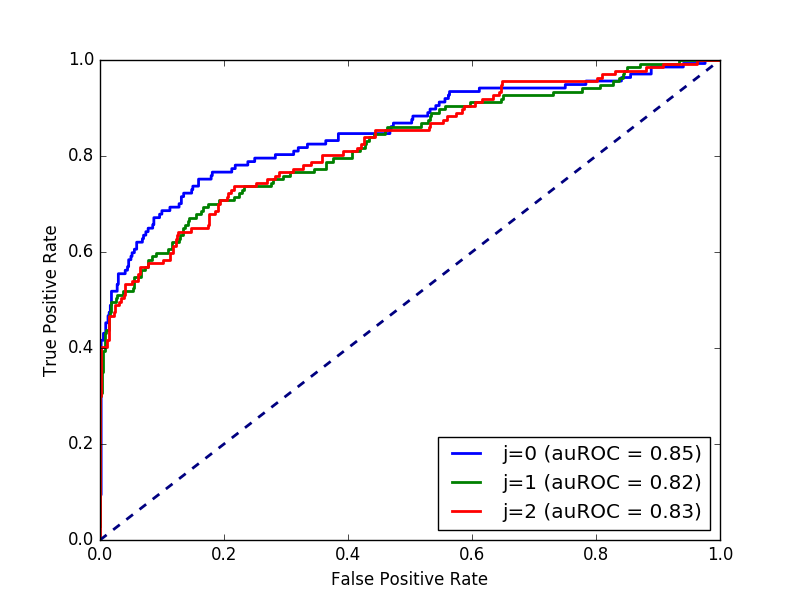}
  \caption{Comparison among different choices of $j\in\{0,1,2\}$ for space-to-channel encoding.}
  \label{fig:seqeunce}
\end{figure}

We construct 3 channel images to AlexNet (late fusion w/ max pooling) by applying $j\in\{0,1,2\}$ in~\eqref{selection} and show the results in Figure~\ref{fig:seqeunce}. We found no significant difference between using single z-slices vs sequences of adjacent z-slices. In fact, performance decreased slightly. We believe this is partially explained by the difference in domain between ImageNet and tomosynthesis z-slices. Natural images such as those found in image net are composed of 3 unique, but structurally similar color channels. Fusing 3 different z-slices produces an image very dissimilar to the expected input, so the extracted features may have less semantic meaning for classification. This issue could potentially be mended by training our network end-to-end initializing AlexNet to the ImageNet pre-trained weights, however for the scope of this research this was not feasible due to computational constraints.

\section{Conclusions}
We propose a novel approach for full-volume DBT classification that enables us to use pre-trained 2D CNNs. In several convincing experiments, our results show significant improvements over the previous 3D CNN based approach. We compare the effects of different pooling methods, feature extractors, and data fusion strategies and find that the combination of \textit{AlexNet}, \textit{max pooling}, and \textit{late fusion strategy} achieves the best performance. We demonstrate the importance of slice-to-slice correspondence information to the classifier and discuss the performance of space-to-channel encoding. These results support the natural assumption that the tissue separation available in DBT z-slices is an invaluable asset in diagnosis. We hope this work is able to provide a baseline as well as a guideline for the future 2D/3D breast cancer image classification research.

\bibliographystyle{IEEEtran}
\bibliography{IEEEfull.bib}

\vspace{12pt}

\end{document}